\pdfoutput=1

\documentclass[11pt]{article}

\usepackage[final]{acl}

\usepackage{times}
\usepackage{latexsym}

\usepackage[T1]{fontenc}

\usepackage[utf8]{inputenc}

\usepackage{microtype}

\usepackage{inconsolata}

\usepackage{graphicx}
\usepackage{booktabs}
\usepackage{pifont}
\usepackage{xcolor}
\usepackage{multirow}
\usepackage{pifont}
\usepackage{float}
\definecolor{sgreen}{RGB}{30, 150, 30} 
\definecolor{green}{RGB}{0,210,0}
\definecolor{blue}{RGB}{218,232,252}
\definecolor{mygray}{RGB}{220,220,220}
\usepackage{graphicx} 
\usepackage{amsmath} 

\usepackage{wrapfig}
\usepackage{float}
\usepackage{colortbl}

\title{Video-LLaVA: Learning United Visual Representation by Alignment Before Projection}

\author{
 \textbf{Bin Lin\textsuperscript{1}},
 \textbf{Yang Ye\textsuperscript{1}},
 \textbf{Bin Zhu\textsuperscript{1}},
 \textbf{Jiaxi Cui\textsuperscript{4}},
\\
 \textbf{Munang Ning\textsuperscript{1,2,3}},
 \textbf{Peng Jin\textsuperscript{1,2,3}},
 \textbf{Li Yuan\textsuperscript{1,2,3}}
\\
\\
 \textsuperscript{1}Peking University Shenzhen Graduate School,
 \textsuperscript{2}Peng Cheng Laboratory,
\\
 \textsuperscript{3}AI for Science (AI4S)-Preferred Program, Peking University Shenzhen Graduate School,
\\
 \textsuperscript{4}PandaVilla Tech Limited
\\
 \small{
   \textbf{Correspondence:} \href{yuanli-ece@pku.edu.cn}{yuanli-ece@pku.edu.cn}
 }
\\
 \small{
   \textbf{GitHub:} \href{https://github.com/PKU-YuanGroup/Video-LLaVA}{https://github.com/PKU-YuanGroup/Video-LLaVA}
 }
}

\begin{document}
\maketitle
\begin{abstract}
Large Vision-Language Model (LVLM) has enhanced the performance of various downstream tasks in visual-language understanding. Most existing approaches encode images and videos into separate feature spaces, which are then fed as inputs to large language models. However, due to the lack of unified tokenization for images and videos, namely misalignment before projection, it becomes challenging for a Large Language Model (LLM) to learn multi-modal interactions from several poor projection layers. 
In this work, we unify visual representation into the language feature space to advance the foundational LLM towards a unified LVLM. As a result, we establish a simple but robust LVLM baseline, \textbf{Video-LLaVA}, which learns from a mixed dataset of images and videos, mutually enhancing each other.
As a result, Video-LLaVA outperforms Video-ChatGPT by \textbf{5.8\%, 9.9\%, 18.6\%, and 10.1\%} on MSRVTT, MSVD, TGIF, and ActivityNet, respectively. Additionally, our Video-LLaVA also achieves superior performances on a broad range of 9 image benchmarks.
Notably, extensive experiments demonstrate that Video-LLaVA mutually benefits images and videos within a unified visual representation, outperforming models designed specifically for images or videos. We aim for this work to provide modest insights into the multi-modal inputs for the LLM.
\end{abstract}

\section{Introduction}

\label{sec:intro}
\begin{figure}[h]
\centering
    \includegraphics[width=1.0\linewidth]{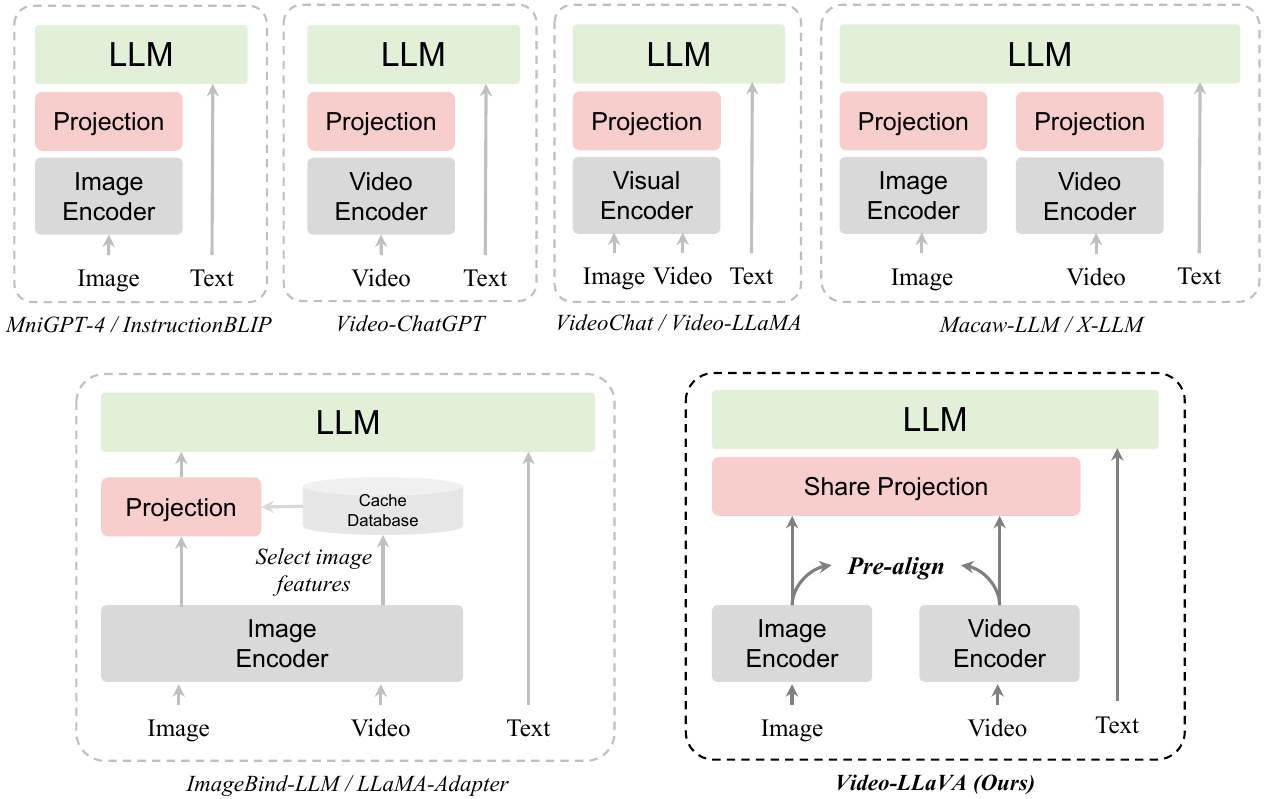}     
\caption{\textbf{Comparing Different LVLM Paradigms.} Video-LLaVA aligns images and videos before projection, allowing LLM to learn from a unified visual representation and endowing LLM with the ability to comprehend both images and videos simultaneously.}
\label{fig:sota}
\end{figure}

Recently, LLMs have gained rapid popularity in the AI community, such as GPT-3.5, GPT-4~\cite{openai2023gpt4}, PaLM~\cite{bi2020palm,anil2023palm}, and BLOOM~\cite{scao2022bloom}. They rely on their powerful language comprehension abilities to follow human-provided instructions and provide corresponding responses. Typically, LLMs can only respond within the text input provided by the user, which is insufficient because human interaction with the world involves multiple channels, such as visual and textual. To this end, recent works~\cite{ye2023mplug,zhu2023minigpt,alayrac2022flamingo} have mapped images into text-like tokens, enabling LLMs to emerge with the ability to comprehend images. Despite their effectiveness, empowering LLMs to understand videos is more challenging than image-only comprehension tasks. Nevertheless, recent work~\cite{maaz2023video,li2023videochat,zhang2023video} has made initial strides in enabling interactions between video and language.

However, most current LVLMs~\cite{li2023blip,dai2023instructblip,luo2023valley,li2023otter,yin2023survey,fu2023mme} can primarily handle a single visual modality, either image-language or video-language. We compare different LVLM paradigms as shown in Figure~\ref{fig:sota}, where VideoChat~\cite{li2023videochat} and Video-LLaMA~\cite{zhang2023video} utilize a share visual encoder to handle both images and videos. However, due to the inherent differences in the media types of images and videos, it is challenging to learn a unified representation, and the performance falls significantly behind that of the specialized video expert model, Video-ChatGPT. Therefore, X-LLM~\cite{chen2023x} and Macaw-LLM~\cite{lyu2023macaw} allocate a modality-specific encoder for each modality, attempting to enable a LLM to comprehend images or videos through several projection layers. But their performances are inferior to dedicated video expert models such as Video-ChatGPT~\cite{maaz2023video}. We attribute this phenomenon to the lack of \textit{\textbf{alignment before projection}}. Because image features and video features reside in their own spaces, this poses a challenge for a LLM to learn their interactions from several poor projection layers. Some similar phenomenon such as \textit{alignment before fusion} has been discussed by ALBEF~\cite{li2021align} and ViLT~\cite{kim2021vilt} in multi-model models. More recently, ImageBind-LLM~\cite{han2023imagebind} focuses on enabling the LLM to simultaneously process multiple modal inputs by pre-aligning each modality to a common feature space~\cite{girdhar2023imagebind}. Based on a large image-language model, ImageBind-LLM converts other modalities into the most similar image features by retrieving from a training-free image cached database. However, the indirect alignment approach of ImageBind-LLM may lead to performance degradation, and the LLM has no knowledge of actual video data.

In this work, we introduce \textbf{Video-LLaVA}, a simple but powerful baseline for the LVLM simultaneously handling both images and videos. Specifically, As shown in Figure~\ref{fig:sota}, Video-LLaVA initially aligns the representations of images and videos to a unified visual feature space. Since the visual representations are already aligned prior to projection, we employ a shared projection layer to map the unified visual representation for the LLM. To enhance computational efficiency, Video-LLaVA undergoes joint training of images and videos, achieving remarkable results with 1 training epoch. 

As a result, The proposed Video-LLaVA greatly enhances the ability of the LLM to simultaneously understand both images and videos. For image understanding, Video-LLaVA surpasses advanced LVLMs such as mPLUG-owl-7B and InstructBLIP-7B in 5 image benchmarks. Additionally, utilizing 4 benchmark toolkits for a more comprehensive evaluation, Video-LLaVA-7B even outperforms IDEFICS-80B by 6.4\% in MMBench. Moreover, similar trends can be observed in video understanding, where Video-LLaVA surpasses Video-ChatGPT by 5.8\%, 9.9\%, 18.6\%, and 10.1\% respectively on the MSVD, MSRVTT, TGIF, and ActivityNet video question-answering datasets. Extensive ablation experiments demonstrate that alignment before projection yields greater benefits. Additionally, joint training of images and videos can facilitate a unified visual representation in LLM comprehension.

We summarize our primary contributions as follows:
\begin{itemize}
\item We introduce \textbf{Video-LLaVA}, a powerful LVLM baseline. During the training process, Video-LLaVA binds visual signals to the language feature space, unifying visual representations, and proposes a solution to align before projection. We enable an LLM to perform visual reasoning capabilities on both images and videos simultaneously.
\item Extensive experiments demonstrate that a unified visual representation benefits LLMs in learning to simultaneously handle both images and videos, validating the complementarity of modalities, showcasing significant superiority when compared to models specifically designed for either images or videos.
\end{itemize}

\section{Related Work}
\label{sec:related}

\subsection{Large Language Models}

When the well-known commercial model ChatGPT~\cite{openai2023gpt4} was introduced, the The AI community released open-source Large Language Models (LLMs) by instruction tuning and increasing model sizes. These include LLaMA~\cite{touvron2023llama}, Vicuna~\cite{chiang2023vicuna}, Alpaca~\cite{taori2023stanford}, and more recently, LLaMA 2~\cite{touvron2023llama2}. These models are tuned with instruction sets to emulate conversations between humans and AI assistants. Furthermore, InstructGPT~\cite{ouyang2022training} is trained based on GPT-3~\cite{brown2020language} with 175 billion parameters through aligning with human preferences. However, LLMs can only interact within text. In this work, we introduce Video-LLaVA, which builds upon the powerful reasoning capabilities of LLM to extend modality interactions to images and videos.

\begin{table*}
  \setlength\tabcolsep{2.0mm}
  \caption{\textbf{Comparison between different Large Vision-Language Models.} For methods that treat LLMs as scheduler, they do not require pre-alignment and joint training.}
  \label{tab:lvlm}
  \centering
  \begin{tabular}{lcccc}
    \toprule
    \textbf{Methods} & \textbf{Image} & \textbf{Video} & \textbf{Pre-aligned} & \textbf{Joint training} \\
    \midrule
    \multicolumn{3}{l}{\textit{LLMs as scheduler}} \\ 
    VisualChatGPT~\cite{wu2023visual} & \textcolor{green}{\ding{52}} & \textcolor{red}{\ding{55}} & - & - \\
    HuggingGPT~\cite{shen2023hugginggpt} & \textcolor{green}{\ding{52}} & \textcolor{red}{\ding{55}} & - & - \\
    MM-REACT~\cite{yang2023mm} & \textcolor{green}{\ding{52}} & \textcolor{green}{\ding{52}} & - & - \\
    ViperGPT~\cite{suris2023vipergpt} & \textcolor{green}{\ding{52}} & \textcolor{green}{\ding{52}} & - & - \\
    \midrule
    \multicolumn{3}{l}{\textit{LLMs as decoder}} \\ 
    Mini-GPT4~\cite{zhu2023minigpt} & \textcolor{green}{\ding{52}} & \textcolor{red}{\ding{55}} & - & \textcolor{red}{\ding{55}} \\
    LLaVA~\cite{liu2023visual} & \textcolor{green}{\ding{52}} & \textcolor{red}{\ding{55}} & - & \textcolor{red}{\ding{55}} \\
    Video-ChatGPT~\cite{maaz2023video} & \textcolor{red}{\ding{55}} & \textcolor{green}{\ding{52}} & - & \textcolor{red}{\ding{55}} \\
    VideoChat~\cite{li2023videochat} & \textcolor{green}{\ding{52}} & \textcolor{green}{\ding{52}} & \textcolor{red}{\ding{55}} & \textcolor{green}{\ding{52}} \\
    Video-LLaMA~\cite{zhang2023video} & \textcolor{green}{\ding{52}} & \textcolor{green}{\ding{52}} & \textcolor{red}{\ding{55}} & \textcolor{green}{\ding{52}} \\
    ImageBind-LLM~\cite{han2023imagebind} & \textcolor{green}{\ding{52}} & \textcolor{green}{\ding{52}} & \textcolor{green}{\ding{52}} & \textcolor{red}{\ding{55}} \\
    \midrule
    \rowcolor{blue} \textbf{Video-LLaVA (Ours)} & \textcolor{green}{\ding{52}} & \textcolor{green}{\ding{52}} & \textcolor{green}{\ding{52}} & \textcolor{green}{\ding{52}} \\
    \bottomrule
  \end{tabular}
\end{table*}

\subsection{Large Vision-Language Models}
When extending LLMs to multi-modal, especially involving images and videos, the main approaches can be categorized into two types in Table~\ref{tab:lvlm}: \textit{i)} treating LLM as a scheduler, \textit{ii)} treating LLM as a decoder.

\subsubsection{LLMs as scheduler} In the scheduler-based methods, various visual models are treated as plug-and-play modules. LLM schedules them according to the specific visual task requirements, like the assembly of building blocks. Some of these methods focus on images, such as VisualChatGPT~\cite{wu2023visual} and HuggingGPT~\cite{shen2023hugginggpt}, while MM-REACT~\cite{yang2023mm} and ViperGPT~\cite{suris2023vipergpt} can also handle videos. A key characteristic of these scheduler-based LVLMs is that they do not require end-to-end training, hence eliminating the need for pre-alignment and joint training of each modality. 

\subsubsection{LLMs as decoder} Regarding the approach of treating LLM as a decoder, this is our primary focus. MiniGPT-4~\cite{zhu2023minigpt} aligns image tokens to the input of the large language model through several linear projection layers. However, this alignment is weak and lacks feedback from human instructions. Subsequently, mPLUG-Owl~\cite{ye2023mplug} adopts a two-stage training approach. In the first stage, images are aligned with language using an auto-regressive pretraining style, and the second stage involves instruction tuning through using a human instruction dataset. With the increasing scale of large language model backends, approaches such as InstructBLIP~\cite{dai2023instructblip} and LLaVA series~\cite{liu2023visual,liu2023improved, lin2024moe} collecte the larger human instruction datasets to train a larger LVLMs (13B parameters). Each answer of instruction datasets strictly follow to the given instructions. Then they undergo end-to-end training using human instruction datasets, enabling the LLM with visual reasoning capabilities. Moreover, Video-ChatGPT~\cite{maaz2023video} design a 100k video instruction dataset, successfully empowering LLMs to comprehend videos. VideoChat~\cite{li2023videochat} and Video-LLaMA~\cite{zhang2023video} achieve this by conducting joint training, allowing LLMs to simultaneously handle images and videos. Expanding LLMs to additional visual modalities typically requires pre-alignment, as seen in LLaMA-Adapter~\cite{zhang2023llama,gao2023llama} and ImageBind-LLM~\cite{han2023imagebind}. They bind other modalities to the image space through ImageBind's~\cite{girdhar2023imagebind} modality encoder. These models have demonstrated that a unified feature space is advantageous for enhancing LLM's multi-modal reasoning capabilities. Distinguished from prior work, Video-LLaVA not only pre-aligns image and video features but also conducts joint training of images and videos, facilitating LLMs in learning multi-modal reasoning capabilities from a unified visual representation.

\section{Video-LLaVA}
\label{sec:videollava}

\subsection{Model Structure}

\subsubsection{Framework Overview}
\begin{figure*}[t]
\centering
    \includegraphics[width=1.0\linewidth]{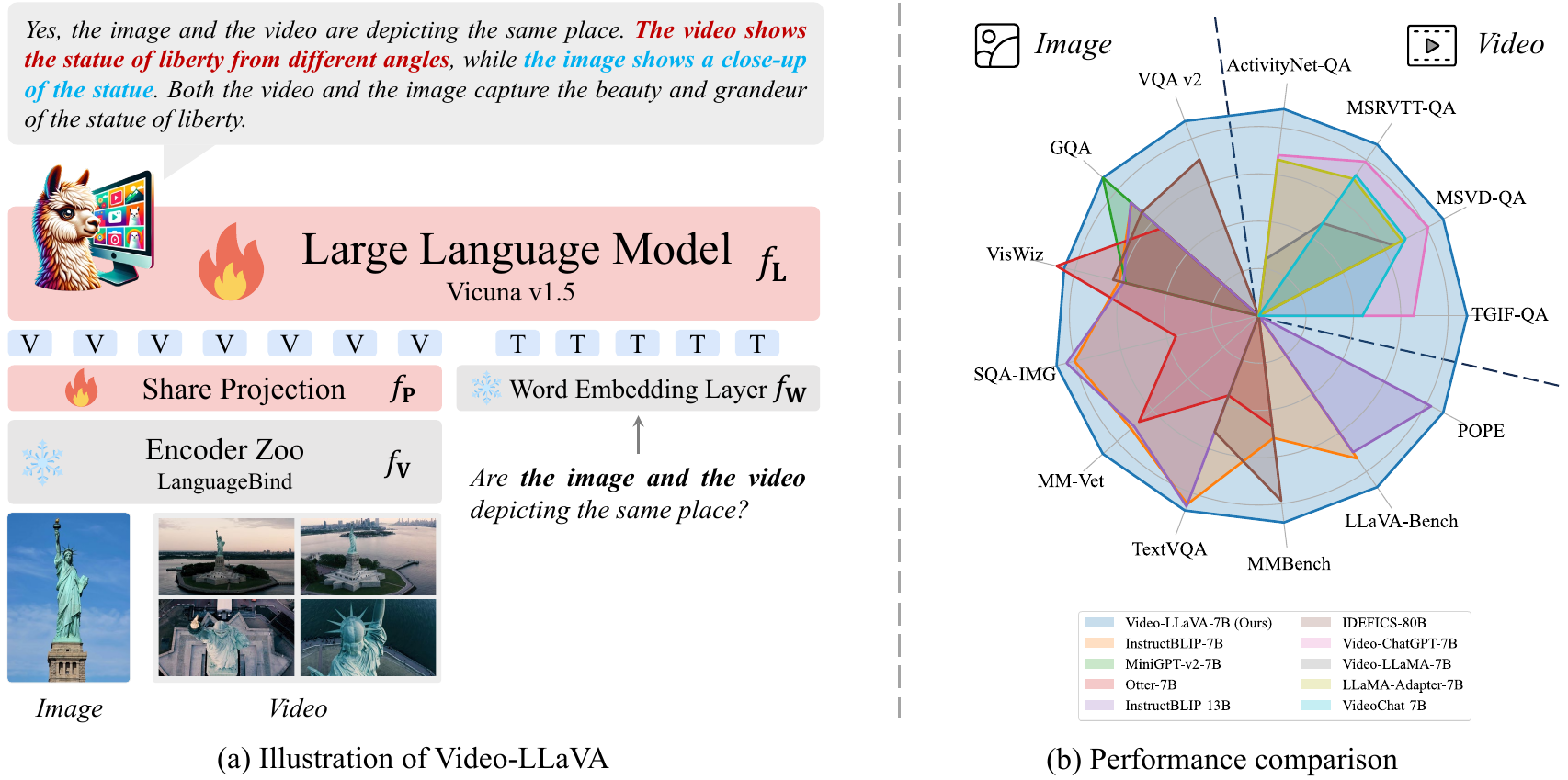}     
\caption{\textbf{Training framework and performance.} Video-LLaVA exhibits remarkable interactive capabilities between images and videos, despite the absence of image-video pairs in the dataset. (a) The Video-LLaVA framework demonstrates a data flow that generates corresponding responses based on input instructions. (b) Video-LLaVA achieves superior performances on a broad range of 15 datasets across image and video.}
\label{fig:videollava}
\end{figure*}

As shown in Figure~\ref{fig:videollava}, Video-LLaVA consists of LanguageBind encoders $f_{\mathbf{V}}$~\cite{zhu2023languagebind} to extract features from the raw visual signal (images or videos), a large language model $f_{\mathbf{L}}$ such as Vicuna, visual projection layers $f_{\mathbf{P}}$ and a word embedding layer $f_{\mathbf{T}}$. We initially obtain visual features using LanguageBind encoders. LanguageBind encoders are capable of mapping different modalities into the textual feature space, thereby providing us with a unified visual representation. Subsequently, the unified visual representation is encoded by shared projection layers, which is then combined with tokenized textual queries and fed into a large language model to generate corresponding responses.

\subsubsection{United Visual Representation}
Our goal is to map images and videos into a shared feature space to enable the large language model to learn from a unified visual representation. We assume that the same information can be conveyed through multiple media. For example, \texttt{a running dog} can be expressed through language, a image or a video simultaneously. Therefore, we can compress information from different modalities into a common feature space, allowing the model to extract information from a dense feature space, facilitating modality interactions and complementarity. Hence, we chose the modality encoders from LanguageBind~\cite{zhu2023languagebind}, which align images and videos with the textual feature space.

\subsubsection{Alignment Before Projection}
Specifically, LanguageBind initializes from OpenCLIP~\cite{ilharco_gabriel_2021_5143773}, naturally aligning images and language in a shared feature space. Subsequently, it aligns video representations to the language space using 3 million video-text pairs from VIDAL-10M~\cite{zhu2023languagebind}. By sharing a language feature space, the image and video representations ultimately converge into a unified visual feature space, which we refer to as emergent alignment of images and videos. Therefore, our video encoder and image encoder are initialized from the LanguageBind encoders zoo, pre-aligning the inputs for LLM and reducing the gap between representations of different visual signals. The unified visual representation is fed into LLM after passing through a shared projection layer.

\subsection{Training Pipeline}
Overall, the process of generating responses by Video-LLaVA is similar to that of a large language model (GPT series). Given a textual input $\mathbf{X}_{\text{T}}$ and visual signals $\mathbf{X}_{\text{V}}$, the input signals are encoded into a sequence of tokens according to Equation~\ref{eq:enc}. By maximizing the likelihood probability in Equation~\ref{eq:gen}, the model ultimately achieves multi-modal understanding capabilities.

\begin{equation}
    \mathbf{Z}_{\text{T}}=f_{\mathbf{T}}\left(\mathbf{X}_{\text{T}}\right), 
    \mathbf{Z}_{\text{V}}=f_{\mathbf{P}}\left(f_{\mathbf{V}}\left(\mathbf{X}_{\text{V}}\right)\right)
  \label{eq:enc}
\end{equation}
\begin{equation}
  p\left(\mathbf{X}_{\text{A}} \mid \mathbf{X}_{\text{V}},\mathbf{X}_{\text{T}}\right)=\prod_{i=1}^L p_\theta\left(\mathbf{X}_{\text{A}}^{[i]} \mid \mathbf{Z}_{\text{V}}, \mathbf{Z}_{\text{T}}^{[1: i-1]}\right)
  \label{eq:gen}
\end{equation}
where $L$ is the length of the generated sequence $\mathbf{X}_{\text{A}}$, and $\theta$ is a trainable parameter. We dynamically conduct joint training on images and videos, wherein a single batch contains both image and video samples simultaneously.

\subsubsection{Understanding Training}
At this stage, the model is required to acquire the ability to interpret visual signals within an extensive image/video-text pair dataset. Each visual signal corresponds to a single round of conversation data $(\mathbf{X}_{\mathrm{q}}, \mathbf{X}_{\mathrm{a}})$, where $\mathbf{X}_{\text{T}}=\mathbf{X}_{\mathrm{q}}$ and $\mathbf{X}_{\mathrm{a}}$ is the ground truth. The training objective of this stage is the original auto-regressive loss, where the model learns the basic ability to view the vision. We freeze the other parameters of the model during this process.

\subsubsection{Instruction Tuning}
In this stage, the model is required to provide responses corresponding to different instructions. These instructions often involve more complex visual comprehension tasks, rather than just describing visual signals. Note that the conversation data $\left(\mathbf{X}_{\mathrm{q}}^1, \mathbf{X}_{\mathrm{a}}^1, \cdots, \mathbf{X}_{\mathrm{q}}^N, \mathbf{X}_{\mathrm{a}}^N\right)$ consists of multiple rounds.
\begin{equation}
\mathbf{X}_{\text{T}}^r=\left\{\begin{array}{lr}
\mathbf{X}_{\mathrm{q}}^1, & r=1 \\
\text{Concat}(\mathbf{X}_{\mathrm{q}}^{r-1}, \mathbf{X}_{\text{A}}^{r-1}, \mathbf{X}_{\mathrm{q}}^r), & r>1
\end{array}\right.
  \label{eq:tuning}
\end{equation}
where $r$ represents the round number. As shown in Equation~\ref{eq:tuning}, when $r>1$ we concatenate the conversations from all previous rounds with the current instruction as the input for this round. The training objective remains the same as in the previous stage. After this stage, the model learns to generate corresponding responses based on different instructions and requests. The LLM are also involved in training at this stage.

\section{Experiments}
\label{sec:exp}

\subsection{Experimental Setup}

\begin{figure}
\vspace{-0.4cm} 
\centering
    \includegraphics[width=1.0\linewidth]{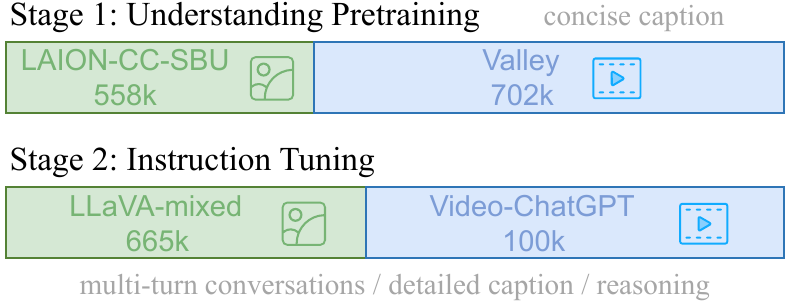}     
    \caption{\textbf{Data composition for training Video-LLaVA.} The dataset for stage 1 consists of single-turn conversation, focusing on concise visual descriptions. In stage 2, the dataset comprises multi-turn conversations, emphasizing complex visual reasoning abilities.}
    \label{fig:data}
\vspace{-0.4cm} 
\end{figure}

\subsubsection{Data Details} In \ref{fig:data}, for the first stage of understanding pretraining, we use a subset of 558K LAION-CC-SBU image-text pairs with BLIP~\cite{li2022blip} captions, which is sourced from CC3M~\cite{sharma2018conceptual} and filtered by LLaVA~\cite{liu2023visual}. The video-text pairs are derived from a subset provided by Valley~\cite{luo2023valley}, and we have access to 702k out of a total of 703k pairs, originating from WebVid~\cite{bain2021frozen}. For the stage of instruction tuning, We gathered instructional datasets from two sources, including a 665k image-text instruction dataset from LLaVA 1.5~\cite{liu2023improved} and a 100k video-text instruction dataset from Video-ChatGPT~\cite{maaz2023video}.

\subsubsection{Model Settings} We employ Vicuna-7B v1.5 as the large language model. The visual encoders are derived from LanguageBind, initialized from OpenCLIP-L/14. The text tokenizer is sourced from LLaMA, with approximately 32,000 classes. The share projection layers consist of 2 fully connected layers with a GeLU~\cite{hendrycks2016gaussian} activated function.

\subsubsection{Training Details} In the training process, we resize and crop each image, resulting in a size of 224×224 for each processed image. We uniformly sample 8 frames from each video, and each frame undergoes image pre-processing. The data in each batch is a random combination of images and videos. In the first stage, we train for one epoch with a batch size of 256, using the AdamW optimizer with a cosine learning rate schedule. In the second stage, we reduce the batch size to 128. The initial learning rate for both stages is set to 1e-3, with a warmup ratio of 0.03. Additional hyper-parameter settings can be found in the appendix.

\subsection{Quantitative Evaluation}

\begin{table*}[t]
  \setlength\tabcolsep{1.55mm}
  \caption{\textbf{Comparison between different LVLMs on video reasoning benchmarks}. We employ ChatGPT-Assistant to evaluate the performance following Video-ChatGPT~\cite{maaz2023video}. The version of ChatGPT is ``gpt-3.5-turbo''.}
  \label{tab:video_qa}
  \centering
  \begin{tabular}{lc|cc|cc|cc|cc}
    \toprule
    \multirow{2}{*}{\textbf{Methods}} & \multirow{1}{*}{\textbf{LLM}} & \multicolumn{2}{c|}{\textbf{MSVD-QA}} & \multicolumn{2}{c|}{\textbf{MSRVTT-QA}} & \multicolumn{2}{c|}{\textbf{TGIF-QA}} & \multicolumn{2}{c}{\textbf{ActivityNet-QA}} \\
     & \multirow{1}{*}{\textbf{size}} & Accuracy & Score & Accuracy & Score & Accuracy & Score & Accuracy & Score \\
     \midrule
    FrozenBiLM & 1B & 32.2 & - & 16.8 & - & 41.0 & - & 24.7 & - \\
    VideoChat & 7B & 56.3 & 2.8 & 45.0 & 2.5 & 34.4 & 2.3 & - & 2.2 \\
    LLaMA-Adapter & 7B & 54.9 & 3.1 & 43.8 & 2.7 & - & - & 34.2 & 2.7 \\
    Video-LLaMA & 7B & 51.6 & 2.5 & 29.6 & 1.8 & - & - & 12.4 & 1.1 \\
    Video-ChatGPT & 7B & 64.9 & 3.3 & 49.3 & 2.8 & 51.4 & 3.0 & 35.2 & 2.7 \\
    Chat-UniVi & 7B & \underline{65.0} & \underline{3.6} & \underline{54.6} & \underline{3.1} & \underline{60.3} & \underline{3.4} & \textbf{45.8} & \underline{3.2} \\
    \rowcolor{blue} \textbf{Video-LLaVA} & \textbf{7B} & \textbf{70.7} & \textbf{3.9} & \textbf{59.2} & \textbf{3.5} & \textbf{70.0} & \textbf{4.0} & \underline{45.3} & \textbf{3.3} \\
    \bottomrule
  \end{tabular}
\end{table*}

\begin{table*}[t]
  \setlength\tabcolsep{1.0mm}
  \caption{\textbf{Comparison between different LVLMs on image understanding benchmarks.} ``Res.'', ``L'', ``V'' respectively represent the input image resolution, LLaMA~\cite{touvron2023llama} and Vicuna~\cite{chiang2023vicuna}. Benchmark names are abbreviated due to page limitations. VQA-v2~\cite{goyal2017making}; GQA~\cite{hudson2019gqa}; VisWiz~\cite{gurari2018vizwiz}; SQA$^\text{I}$: ScienceQA-IMG~\cite{lu2022learn}; VQA$^\text{T}$: TextVQA~\cite{singh2019towards}; POPE~\cite{li2023evaluating}; MMB: MMBench~\cite{liu2023mmbench}; LLaVA$^\text{W}$: LLaVA-Bench (In-the-Wild)~\cite{liu2023visual}; MM-Vet~\cite{yu2023mm}. $^\dag$ donates that we reproduce LLaVA-1.5 with LanguageBind-Image encoder to compare fairly. $^*$ donates that there is some overlap in the training data.}
  \label{tab:image_res}
  \centering
  \begin{tabular}{llc|ccccc|cccc}
    \toprule
     \multirow{2}{*}{\textbf{Methods}} & \multirow{2}{*}{\textbf{LLM}} & \multirow{2}{*}{\textbf{Res.}} & \multicolumn{5}{c|}{\textbf{Image Question Answering}} & \multicolumn{4}{c}{\textbf{Benchmark Toolkit}} \\
      &  &  & VQA$^\text{v2}$ & GQA & VisWiz & SQA$^\text{I}$ & VQA$^\text{T}$ & POPE & MMB & LLaVA$^\text{W}$ & MM-Vet \\
    \midrule
    \rowcolor{mygray} \color{gray} LLaVA-1.5 & \color{gray} V-7B & \color{gray} 336 & \color{gray} - & \color{gray} 62.0$^*$ & \color{gray} - & \color{gray} - & \color{gray} - & \color{gray} - & \color{gray} - & \color{gray} - & \color{gray} 30.5 \\
    \rowcolor{mygray} \color{gray} BLIP-2 & \color{gray} V-13B & \color{gray} 224 & \color{gray} 41.0 & \color{gray} 41.0 & \color{gray} 19.6 & \color{gray} 61.0 & \color{gray} 42.5 & \color{gray} 85.3 & \color{gray} - & \color{gray} 38.1 & \color{gray} 22.4 \\
    \rowcolor{mygray} \color{gray} InstructBLIP & \color{gray} V-13B & \color{gray} 224 & \color{gray} - & \color{gray} 49.5 & \color{gray} 33.4 & \color{gray} 63.1 & \color{gray} 50.7 & \color{gray} 78.9 & \color{gray} - & \color{gray} 58.2 & \color{gray} 25.6 \\
    \rowcolor{mygray} \color{gray} IDEFICS-80B & \color{gray} L-65B & \color{gray} 224 & \color{gray} 60.0 & \color{gray} 45.2 & \color{gray} 36.0 & \color{gray} - & \color{gray} 30.9 & \color{gray} - & \color{gray} 54.5 & \color{gray} - & \color{gray} - \\
    MiniGPT-4 & L-7B & 224 & - & 30.8 & 47.5 & 25.4 & 19.4 & - & 23.0 & - & 22.1 \\
    IDEFICS-9B & L-7B & 224 & {50.9} & 38.4 & 35.5 & - & 25.9 & - & {48.2} & - & - \\
    mPLUG-Owl & L-7B & 224 & - & 14.0 & 39.0 & 2.8 & 38.8 & - & 46.6 & - & - \\
    Otter & L-7B & 224 & - & 38.1 & \textbf{50.0} & 27.2 & 21.2 & - & 32.6 & - & 24.6 \\
    InstructBLIP & V-7B & 224 & - & {49.2} & 34.5 & {60.5} & \underline{50.1} & - & 36.0 & {60.9} & \underline{26.2} \\
    LLaVA-1.5$^{\dag}$ & V-7B & 224 & \underline{72.3}$^*$ & \underline{56.9}$^*$ & 47.8 & \textbf{67.9} & 49.2 & \underline{83.3} & \underline{59.5} & \underline{63.3} & 25.7 \\
    \rowcolor{blue} \textbf{Video-LLaVA} & \textbf{V-7B} & \textbf{224} & \textbf{74.7}$^*$ & \textbf{60.3}$^*$ & \underline{48.1} & \underline{66.4} & \textbf{51.8} & \textbf{84.4} & \textbf{60.9} & \textbf{73.1} & \textbf{32.0} \\
    \bottomrule
  \end{tabular}
\end{table*}

\subsubsection{Zero-shot Video Understanding}
As shown in Table~\ref{tab:video_qa}, we conduct a quantitative assessment of the video question-answering capabilities of large video-language models on four datasets, including MSVD-QA~\cite{chen2011collecting}, MSRVTT-QA~\cite{xu2016msr}, TGIF-QA~\cite{jang2017tgif} and ActivityNet-QA~\cite{yu2019activitynet}. The evaluation pipeline for video understanding follows Video-ChatGPT. We report the accuracy and score, which is assessed using GPT-Assistant. Video-LLaVA consistently outperforms Video-ChatGPT in terms of question-answering accuracy, which is an advanced large video-language model. Moreover, Video-LLaVA surpasses the powerful baseline of Video-ChatGPT by 5.8\%, 9.9\%, 18.6\%, and 10.1\% on MSRVTT, MSVD, TGIF, and ActivityNet, respectively. Additionally, we conduct comparisons with the recent SOTA model, Chat-UniVi~\cite{Chat-UniVi}. Despite Chat-UniVi utilizing more datasets such as MIMIC-IT~\cite{li2023otter}, Video-LLaVA still demonstrate competitive results, surpassing Chat-UniVi on MSVD, MSRVTT, and TGIF datasets. In summary, these results validate Video-LLaVA's ability to comprehend videos and provide contextually appropriate responses based on instructions.

\begin{table*}[t]
  \setlength\tabcolsep{0.65mm}
  \caption{\textbf{Zero-shot object hallucination evaluation results} are reported for three POPE evaluation settings. ``Yes'' indicates the proportion of positive responses to the given question. $^\dag$ donates that we reproduce LLaVA-1.5 with LanguageBind-Image encoder to compare fairly.}
  \label{tab:pope}
  \centering
  \begin{tabular}{ll|ccc|ccc|ccc}
    \toprule
    \multirow{2}{*}{\textbf{Methods}} & \multirow{2}{*}{\textbf{LLM}}& \multicolumn{3}{c|}{\textbf{Adersarial}} & \multicolumn{3}{c|}{\textbf{Popular}}  & \multicolumn{3}{c}{\textbf{Random}}  \\
     &  & Accuracy & F1-Score & Yes & Accuracy & F1-Score & Yes & Accuracy & F1-Score & Yes \\
    \midrule
    \rowcolor{mygray} \color{gray} MiniGPT-4 & \color{gray} V-13B & \color{gray} 66.6 & \color{gray} 71.4 & \color{gray} 66.7 & \color{gray} 68.3 & \color{gray} 72.2 & \color{gray} 64.1 & \color{gray} 77.8 & \color{gray} 78.9 & \color{gray} 54.8 \\
    \rowcolor{mygray} \color{gray} InstructBLIP & \color{gray} V-13B & \color{gray} {74.4} & \color{gray} {78.5} & \color{gray} 69.0 & \color{gray} {81.4} & \color{gray} {83.5} & \color{gray} 62.6 & \color{gray} {88.7} & \color{gray} {89.3} & \color{gray} 55.2 \\
    MM-GPT & L-7B & 50.0 & 66.7 & 100.0 & 50.0 & 66.7 & 100.0 & 50.0 & 66.7 & 100.0 \\
    mPLUG-Owl & L-7B & 50.7 & 66.8 & 98.7 & 50.9 & 66.9 & 98.6 & 54.0 & 66.4 & 95.6 \\
    Chat-UniVi & V-7B & {55.6} & {68.7} & 91.6 & {56.4} & {69.0} & 90.8 & {73.9} & {79.3} & 74.6 \\
    LLaVA-1.5$^\dag$ & L-7B & \textbf{84.3} & \textbf{83.2} & 43.5 & \underline{79.8} & \underline{79.4} & 48.0 & \underline{85.7} & \underline{84.8} & 43.0 \\
    \rowcolor{blue} \textbf{Video-LLaVA} & \textbf{V-7B} & \underline{81.6} & \underline{80.8} & 45.8 & \textbf{85.3} & \textbf{84.0} & 42.1 & \textbf{86.2} & \textbf{85.2} & 42.0 \\
    \bottomrule
  \end{tabular}
\end{table*}

\subsubsection{Zero-shot Image Question-answering} As shown in Table~\ref{tab:image_res}, we evaluate our approach for image understanding on five academic image question-answering benchmarks. Compared to the state-of-the-art model InstructBLIP-7B, Video-LLaVA demonstrates powerful image understanding capabilities, outperforming across all five question-answering benchmarks. Additionally, Video-LLaVA exhibits competitive results compared to several more powerful LVLMs, which are tuned based on 13B or 65B LLM, such as surpassing InstructBLIP-13B by 14.7\% on VisWiz, highlighting its strong understanding ability in natural visual environments. Furthermore, to ensure a fair comparison, we replace the image encoder in LLaVA-1.5 with the LanguageBind-Image encoder, called LLaVA-1.5$^{\dag}$. This demonstrates that the performance improvement observed in Video-LLaVA is not solely attributed to a stronger image encoder. Additional details can be found in Section~\ref{sec:jt_img}.

\vspace{0.1cm}
\noindent\textbf{Evaluation under Image Benchmark Toolkits} Additionally, we evaluate LVLMs using several benchmark toolkits for visual instruction tuning. These benchmark toolkits provide a detailed assessment of the model's capabilities through robust evaluation metrics. Video-LLaVA outperform InstructBLIP-7B by 24.9\%, 12.2\%, and 5.8\% on MMBench, LLaVA-Bench, and MM-Vet, respectively. It is worth noting that Video-LLaVA-7B still demonstrates advanced performance compared to larger LLM models, surpassing InstructBLIP-13B by 6.4\% on MM-Vet and IDEFICS-80B~\cite{laurencon2023obelics} by 6.4\% on MMBench. These results demonstrate that Video-LLaVA exhibits a strong understanding of semantic aspects of scenes, enabling it to answer open-ended and free-form natural language questions about images.

\subsubsection{Object Hallucination Evaluation} As shown in Table~\ref{tab:pope}, we report evaluation results for zero-shot object hallucinations, utilizing a evaluation pipeline derived from a polling-based query method~\cite{li2023evaluating}. Video-LLaVA demonstrates competitive performance across three subsets: random, popular, and adversarial. Specifically, when compared to the 7B foundation model, Video-LLaVA consistently outperforms MM-GPT~\cite{gong2023multimodal} across all three POPE hallucination evaluation subsets. Furthermore, when benchmarked against the larger 13B LLM, Video-LLaVA even surpasses Mini-GPT4 comprehensively. The successful performance of Video-LLaVA in object hallucination detection validates the consistency between unified visual representations and the generation of textual descriptions.

\subsection{Ablation Results}

\subsubsection{Alignment Before Projection} To validate the performance degradation caused by separated visual representation, we conduct experiments to to explore the performance of the LLM learning from different visual representations. We define the use of LanguageBind image encoder as unified visual representation while the MAE encoder~\cite{he2022masked} use separated visual representation, which is a well-known and effective image feature extractor. Additionally, since MAE do not interact with multi-modal inputs during the training process, we utilize CLIP-L/14, a model of the same size. While CLIP-L/14 exhibits strong multimodal understanding capabilities, it is not pre-aligned with the video encoder. Consequently, this results in a lack of uniformity in the visual features provided to LLM. We only replace the image encoder of the same scale and keep the LanguageBind video encoder. 

\begin{table*}[t]
  \setlength\tabcolsep{0.55mm}
  \caption{\textbf{Effect of alignment before projection on image.} ``\textcolor[rgb]{0.12,0.47,0.71}{United}'' refers to the unified visual representation, while ``\textcolor[rgb]{1.0,0.5,0.05}{Separated}'' refers to the separated visual representation. Benchmark names are abbreviated due to page limitations.}
  \label{tab:uvr_fig}
  \centering
  \begin{tabular}{l|ccccc|cccc}
    \toprule
     \multirow{2}{*}{\textbf{Methods}} & \multicolumn{5}{c|}{\textbf{Image Question Answering}} & \multicolumn{4}{c}{\textbf{Benchmark Toolkit}} \\
     & VQA$^\text{v2}$ & GQA & VisWiz & SQA$^\text{I}$ & VQA$^\text{T}$ & POPE & MMB & LLaVA$^\text{W}$ & MM-Vet \\
    \midrule
    \textcolor[rgb]{1.0,0.5,0.05}{Separated}-MAE & 66.0 & 55.4 & 42.5 & 65.0 & 44.2 & \underline{80.8} & 45.7 & 35.9 & 20.0 \\
    \textcolor[rgb]{1.0,0.5,0.05}{Separated}-CLIP & \underline{74.6} & \underline{59.9} & \underline{47.8} & \textbf{67.3} & \underline{51.5} & \textbf{84.4} & \underline{60.2} & \underline{68.9} & \underline{30.6} \\
    \textcolor[rgb]{0.12,0.47,0.71}{United} & \textbf{74.7} & \textbf{60.3} & \textbf{48.1} & \underline{66.4} & \textbf{51.8} & \textbf{84.4} & \textbf{60.9} & \textbf{73.1} & \textbf{32.0} \\
     $\Delta$ \textcolor{sgreen}{$Acc.$} & \textcolor{sgreen}{{\textbf{+0.1}}} & \textcolor{sgreen}{{\textbf{+0.4}}} &\textcolor{sgreen}{{\textbf{+0.3}}} &\textcolor{gray}{{\textbf{-0.9}}} &\textcolor{sgreen}{{\textbf{+0.3}}} &\textcolor{sgreen}{{\textbf{+0.0}}} &\textcolor{sgreen}{{\textbf{+0.7}}} &\textcolor{sgreen}{{\textbf{+4.2}}} &\textcolor{sgreen}{{\textbf{+1.4}}} \\
    \bottomrule
  \end{tabular}
\end{table*}

\begin{table*}[t]
  \setlength\tabcolsep{0.5mm}
  \caption{\textbf{Effect of joint training on video.} We evaluate on four video question-answering datasets. $^*$ denotes that we utilized only video data in both the first and second stages.}
  \label{tab:jl_tab}
  \centering
  \begin{tabular}{l|cc|cc|cc|cc}
    \toprule
    \multirow{2}{*}{\textbf{Methods}} & \multicolumn{2}{c|}{\textbf{MSVD-QA}} & \multicolumn{2}{c|}{\textbf{MSRVTT-QA}} & \multicolumn{2}{c|}{\textbf{TGIF-QA}} & \multicolumn{2}{c}{\textbf{ActivityNet-QA}} \\
     & Accuracy & Score & Accuracy & Score & Accuracy & Score & Accuracy & Score \\
     \midrule
    Video-LLaVA$^*$ & 64.8 & 3.2 & 58.3 & 3.4 & 67.8 & 3.4 & 40.7 & 2.0 \\
    Joint with Image & \textbf{70.7} & \textbf{3.9} & \textbf{59.2} & \textbf{3.5} & \textbf{70.0} & \textbf{4.0} & \textbf{45.3} & \textbf{3.3} \\
    $\Delta$ \textcolor{sgreen}{$Acc.$} & \textcolor{sgreen}{{\textbf{+5.9}}} & \textcolor{sgreen}{{\textbf{+0.7}}} & \textcolor{sgreen}{{\textbf{+0.9}}} & \textcolor{sgreen}{{\textbf{+0.1}}} & \textcolor{sgreen}{{\textbf{+2.2}}} & \textcolor{sgreen}{{\textbf{+0.6}}} & \textcolor{sgreen}{{\textbf{+4.6}}} & \textcolor{sgreen}{{\textbf{+1.3}}} \\
    \bottomrule
  \end{tabular}
\end{table*}

\begin{figure}[t]
\centering
    \includegraphics[width=1.0\linewidth]{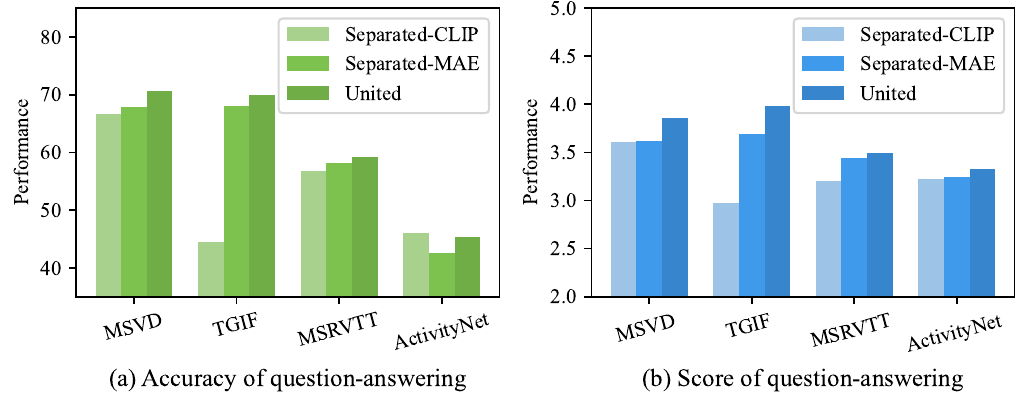}     
  \caption{\textbf{Effect of alignment before projection on video.} We validate and report the accuracy and score on four video question-answering datasets.}
\label{fig:uvr_video}
\end{figure}

\begin{figure}[t]
\centering
    \includegraphics[width=1.0\linewidth]{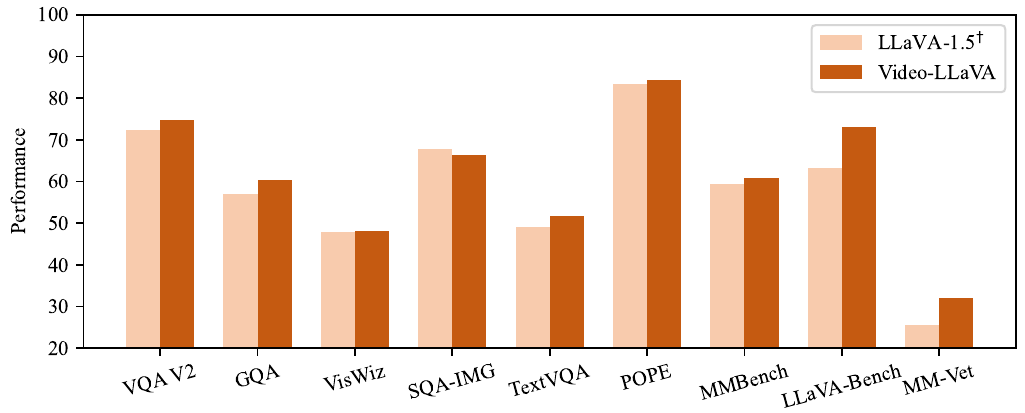}     
\caption{\textbf{Effect of joint training on image.} $^\dag$  donates that We reproduce the results of LLaVA-1.5 at a resolution of 224×224 
 with LanguageBind-Image encoder for a fair comparison.}
\label{fig:jl_fig}
\end{figure}

\subsubsection{For Video Understanding} Due to replacing the image encoder with the MAE encoder, the video features and image features are no longer unified during LLM's initial learning of visual representations. In Figure~\ref{fig:uvr_video}, compared to separated visual representation, the united visual representation significantly improves performance across 4 video question-answering datasets. Separated visual representations not only exhibit lower accuracy in question-answering, but also demonstrate a similar trend in answer scores. These results demonstrate that the unified visual representation can help the LLM further learn and understand videos.

\subsubsection{For Image Understanding} The unified visual representation demonstrates strong performance, surpassing the separated visual representation comprehensively across 5 image question-answering datasets and 4 benchmark toolkits in Table~\ref{tab:uvr_fig}. Additionally, we observe a significant margin of performance improvement in the unified visual representation on the MMBench, LLaVA-Bench, and MM-Vet benchmark toolkits. This highlights that the unified visual representation not only enhances performance in image question-answering but also provides benefits in other aspects of image understanding, such as reducing object hallucination and improving OCR capabilities.

\subsubsection{Joint Training}
This subsection aims to validate the complementarity of images and videos during joint training, which can mutually enhance the LLM's understanding of images and videos based on a unified visual representation. 

\subsubsection{For Video Understanding} For comparing performance on video benchmarks, we remove image data during the training of Video-LLaVA, which is called Video-LLaVA$^*$. We compare with Video-LLaVA$^*$ to assess the performance gains from joint image training on video benchmarks. In Table~\ref{tab:jl_tab}, we evaluate our model on four video question-answering datasets. Compared to Video-LLaVA$^*$ without image in training, the model trained with joint images and videos achieves comprehensive improvements across all four video datasets. These results demonstrate that joint training of images and videos facilitates LLM's understanding of visual representations.

\subsubsection{For Image Understanding}
\label{sec:jt_img}
When comparing performance on image benchmarks, it is challenging to find a image-based LVLM with the same configuration as Video-LLaVA. To address this, we replace the image encoder in LLaVA-1.5 with the LanguageBind-Image encoder and reproduce the results at a resolution of 224×224 by using the same training configuration, called LLaVA-1.5$^\dag$. As shown in Figure~\ref{fig:jl_fig}, Compared to LLaVA-1.5$^\dag$, which utilizes the same image encoder configuration, we observe performance improvements in 8 out of 9 benchmarks, demonstrating mutual improvement in visual understanding. Video-LLaVA outperform LLaVA-1.5$^\dag$ in POPE, indicating that joint training with videos alleviates the object hallucination in images. The similar trend is observed on some other benchmark toolkits, such as LLaVA-Bench and MMBench, where video data significantly improves LLM's performance in complex reasoning and image conversation tasks.

\section{Limitation and Future Directions}

\subsection{Limitation} While Video-LLaVA exhibits strong competitiveness in both images and videos, we still observed some limitations of Video-LLaVA. To begin with, Video-LLaVA performs moderately in understanding long videos. In Table~\ref{tab:video_qa}, Chat-UniVi surpasses 0.5 on ActivityNet-QA because Video-LLaVA only utilizes uniformly sampled 8 frames to comprehend the video, which results in the loss of detailed information from long videos. Additionally, training Video-LLaVA is computationally expensive, requiring 3-4 days to complete the training process on 8 A100-80G GPUs. 

\subsection{Future Directions} In the future, We maybe can explore more efficient shared projection mode that can compress tokens while preserving data features. This would support Video-LLaVA in better understanding long videos. Besides, Video-LLaVA can serve as a baseline to extend to additional visual-related modalities, such as depth and infrared images. Additionally, we could explore how to incorporate timestamp embeddings effectively, enabling large visual-language models to answer questions related to temporal relationships.

\section{Conclusion}
\label{sec:conclu}
In this work, we introduce Video-LLaVA, a simple but powerful large visual-language baseline model. We propose a novel framework to address the issue of misalignment before projection, utilizing a LanguageBind encoder to pre-bind visual signals into the language feature space. To enable a LLM to comprehend both images and videos simultaneously, we conduct joint training on images and videos, allowing the LLM to learn multi-modal interactions from a unified visual representation. Extensive experiments demonstrate that joint training on images and videos mutually benefits performance. Furthermore, we validate that aligning visual representations before projection aids LLM learning. Remarkably, LLM, after learning from a unified visual representation, exhibits the remarkable ability to simultaneously engage with both images and videos, showcasing a powerful comprehension of unified visual concepts. These results collectively demonstrate the effectiveness of the Video-LLaVA training framework. As a unified visual training framework, the performance of Video-LLaVA even surpasses that of expert models designed specifically for images or videos.

\section*{Acknowledgments}

This work was supported in part by the Natural Science Foundation of China (No. 62202014, 62332002, 62425101), Shenzhen Basic Research Program (No.JCYJ20220813151736001).

\clearpage
\bibliography{main}

\clearpage

\appendix

\section{Example Appendix}
\label{sec:appendix}

\subsection{Training Setting}
\label{sec:Setting}

We show some training settings as shown in Table~\ref{tab:setting}. video encoder and image encoder are not trained in both stages. The projection layer consists of 2 linear layers with a GeLU~\cite{hendrycks2016gaussian} activation function between them. Image and video share the projection layer.

\begin{table}[htbp]
\setlength\tabcolsep{1.5mm}
    \small
    \captionof{table}{Training setting.}
    \label{tab:setting}
    \centering
        \begin{tabular}{l|c|c}
            \toprule
            Config & Pretraining & Instruction tuning \\
            \midrule
            Video encoder &  \multicolumn{2}{c}{LanguageBind-Video-LoRA} \\
            Image encoder &  \multicolumn{2}{c}{LanguageBind-Image} \\
            Optimizer & \multicolumn{2}{c}{AdamW} \\
            Deepspeed & \multicolumn{2}{c}{Zero2} \\
            Epochs & \multicolumn{2}{c}{1} \\
            Vision select layer & \multicolumn{2}{c}{-2} \\
            Weight decay & \multicolumn{2}{c}{0.0} \\
            Warmup ratio & \multicolumn{2}{c}{0.03} \\
            Learning rate schedule & \multicolumn{2}{c}{cosine decay} \\
            Learning rate & 1e-3 & 2e-5  \\
            Batch size & 256 & 128 \\
            \bottomrule
        \end{tabular}
\end{table}

\subsection{Exhibition Board}
\label{sec:Exhibition}
We show some \textbf{unselected} samples here, and these videos are sourced from Video-ChatGPT~\cite{maaz2023video}.

\begin{figure}[h]
\centering
\includegraphics[width=0.85\linewidth]{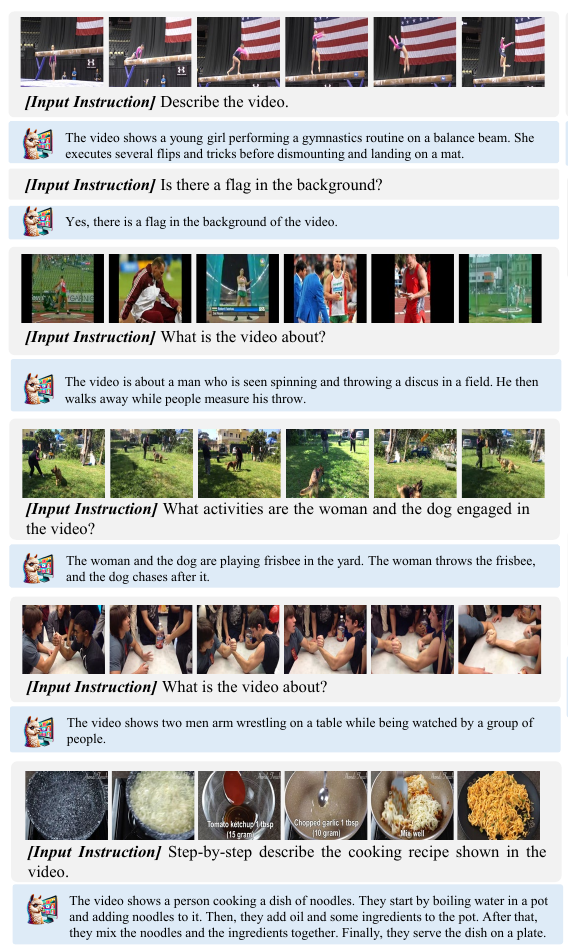}
\caption{Samples of Video-LLaVA in video understanding.}
\label{fig:appdenx1}
\end{figure}

\begin{figure}[h]
\centering
\includegraphics[width=0.65\linewidth]{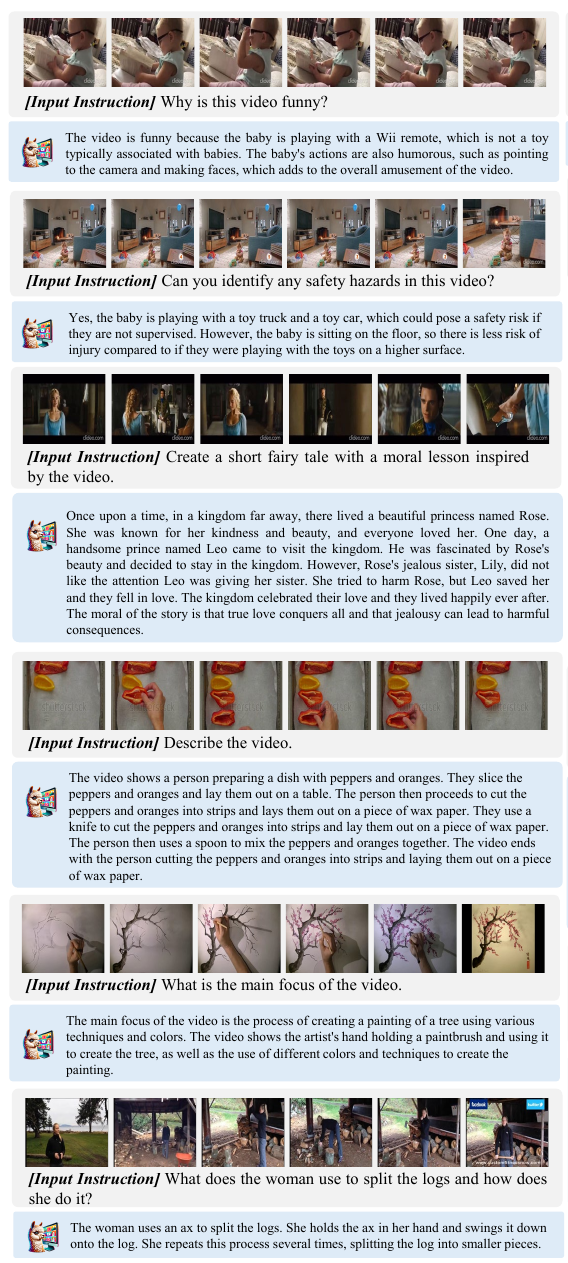}
\caption{Samples of Video-LLaVA in video understanding.}
\label{fig:appdenx3}
\end{figure}

\begin{figure}[h]
\centering
\includegraphics[width=0.65\linewidth]{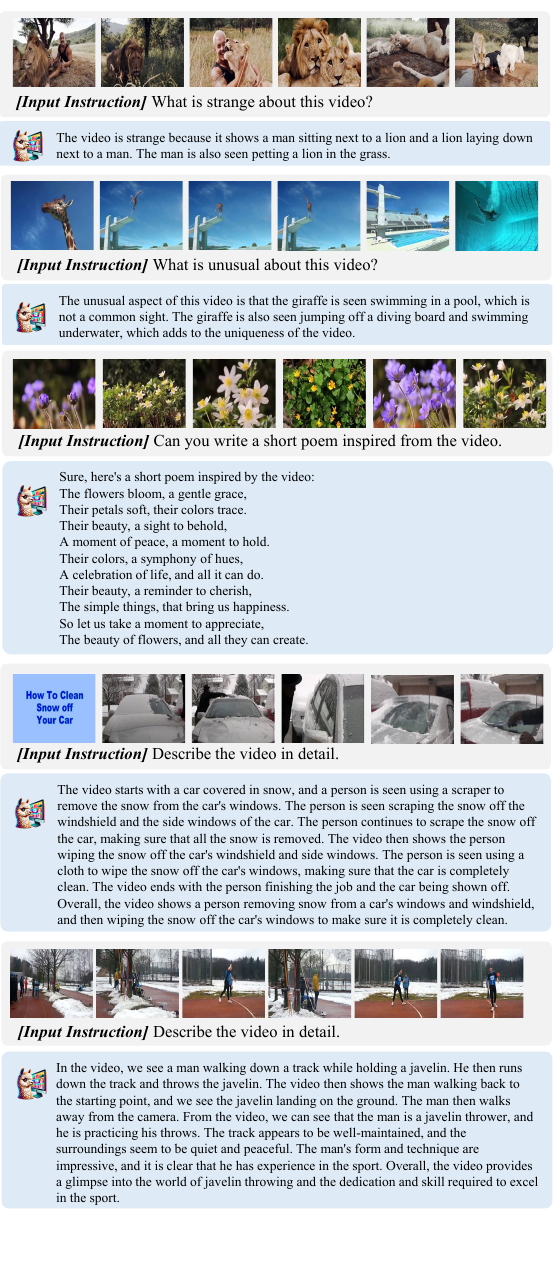}
\caption{Samples of Video-LLaVA in video understanding.}
\label{fig:appdenx4}
\end{figure}

\begin{figure}[h]
\centering
\includegraphics[width=0.85\linewidth]{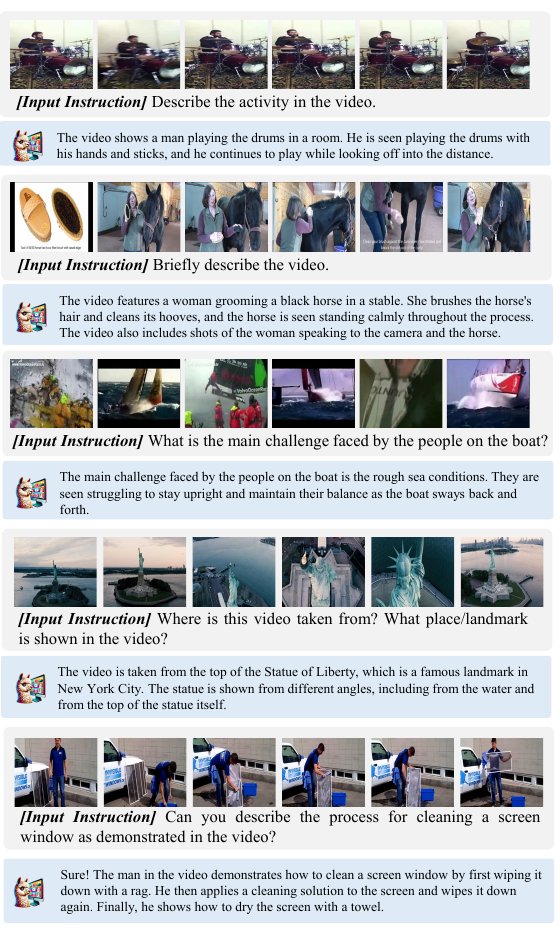}
\caption{Samples of Video-LLaVA in video understanding.}
\label{fig:appdenx2}
\end{figure}

\end{document}